\documentclass[10pt, a4paper]{article}
\usepackage{authblk}
\usepackage{cite}
\usepackage{url}
\usepackage[pass]{geometry}
\usepackage[utf8]{inputenc}
\usepackage[dvipdfmx]{graphicx}
\usepackage{subfig}
\usepackage{ascmac}
\usepackage{bm}
\usepackage{dsfont}
\usepackage{amsmath}
\usepackage{empheq}
\usepackage{comment}
\usepackage{amssymb}
\usepackage{url}
\usepackage{nccmath}
\usepackage{mathtools}
\usepackage{footnote}
\makesavenoteenv{tabular}
\makesavenoteenv{table}
\usepackage[colorlinks=true, bookmarks=true, bookmarksnumbered=true, bookmarkstype=toc, linkcolor=blue, urlcolor=blue, citecolor=blue]{hyperref}

\renewenvironment{abstract}{\nobreak}

\begin{document}

\title{\textbf{Unsupervised Learning of Harmonic Analysis Based on Neural HSMM with Code Quality Templates}}
\author[1]{Yui Uehara}
\affil[1]{Kanagawa University}
\affil[]{yuiuehara@kanagawa-u.ac.jp}
\date{}

\maketitle

\begin{abstract}
\textbf{Abstract.}
This paper presents a method of unsupervised learning of harmonic analysis based on a hidden semi-Markov model (HSMM). We introduce the chord quality templates, which specify the probability of pitch class emissions given a root note and a chord quality. Other probability distributions that comprise the HSMM are automatically learned via unsupervised learning, which has been a challenge in existing research. The results of the harmonic analysis of the proposed model were evaluated using existing labeled data. While our proposed method has yet to perform as well as existing models that used supervised learning and complex rule design, it has the advantage of not requiring expensive labeled data or rule elaboration. Furthermore, we also show how to recognize the tonic without prior knowledge, based on the transition probabilities of the Markov model.
\end{abstract}

\section{Introduction}
\renewcommand{\thefootnote}{\fnsymbol{footnote}}
\footnote[0]{The original edition of this paper will be published in the ICNMC2024 Proceedings and this arXiv publication is a copy.}
\renewcommand{\thefootnote}{\arabic{footnote}}
Harmonic analysis is the process of representing a musical piece as a sequence of chord labels, which facilitates understanding the structure of the piece. In tonal music, the chord label is called chord degree. The chord degree represents chords with a position (denoted by a Roman numeral) on the scale of a local key. This approach to labeling chords in the context of the key is based on the idea that chord progressions play a crucial role in establishing a tonality, which is fundamental in tonal music. Therefore, harmonic analysis can be applied to various tasks such as composition\cite{groves2013,tsushima2017} and higher-level music analysis\cite{rohrmeier2011,wilding2012}.

Several studies have explored automated harmonic analysis\cite{temperley_alg,temperley_cog,temperley_mod,temperley_kru,temperley_web,masada_crf,chen_ht,chen_att,micchi_rome,micchi_deep,raphael_gra,wang2012}. Among them, unsupervised learning is advantageous as it does not require expensive labeled data. However, there have been few works on unsupervised harmonic analysis\cite{raphael_gra,wang2012}. The challenge with harmonic analysis is that it involves simultaneously identifying both keys and chords. Since many possible combinations exist, obtaining an optimal result through unsupervised learning is challenging. Therefore, in order to make the unsupervised learning tractable, previous studies relied on manually designed model parameters\cite{raphael_gra,wang2012}. In this sense, the models were not fully unsupervised.

This study presents a new method for unsupervised harmonic analysis\footnote{\url{https://github.com/yui-u/harmonic-analysis-chorales}}. 
The proposed model is based on a hidden semi-Markov model.
Most model parameters can be learned automatically using non-labeled data, which is a departure from the previous approaches. 
As an exception, code quality templates are set manually.
The chord quality templates correspond to chord labels in supervised learning, which allow comparison with existing harmonic analysis.
We construct the semi-Markov model with the technique of deep latent variable models~\cite{kim2018}.
The technique is to employ neural networks to approximate probability distributions that comprise a targeted latent variable model, which helps unsupervised learning.

Although the experimental results show that our model still has room to be improved, we exemplify the potential of our model with automatic evaluations and discussions on the obtained harmonic analysis.
We also discuss how transition probabilities of the model obtained by unsupervised learning can find the tonic without prior knowledge.

This paper is organized as follows. In Section~\ref{sec:related-work}, we review related studies. We introduce the proposed model in Section~\ref{sec:methodology}. We describe experimental setups and results in Section~\ref{sec:experiments}. Then, we summarize our contributions in Section~\ref{sec:conclusion}.

\section{Related Work}
\label{sec:related-work}
\subsection{Automated harmonic analyses}
\subsubsection{Rule-based harmonic analysis}
The Melisma Music Analyzer is one of the leading harmonic analysis models, comprising the Meter, Grouper, Streamer, Harmony, and Key programs\cite{temperley_alg,temperley_cog,temperley_mod,temperley_kru,temperley_web}. Since the chord labels (Roman numerals) in harmonic analysis identify chords within the context of keys, they require information about both keys and chords. The pipeline analyses using Meter, Harmony, and Key programs can provide Roman numerals. The system is rule-based, and utilizes musical knowledge to determine chord tones. Additionally, it provides various criteria for dealing with ambiguous events, for example, preferring chord changes on strong beats and root progression on the line of fifth (circle of fifth)\cite{temperley_cog}. However, writing down all the preference rules and their priorities to deal with this ambiguity takes much work. This could pose a challenge when expanding the system to cover different types of music.

\subsubsection{Harmonic analyses with supervised learning}
On the other hand, supervised learning relies on high-quality labeled data to learn chord discriminators automatically. Masada and Bunescu employed a semi-Markov Conditional Random Field (semi-CRF) for supervised harmonic analysis\cite{masada_crf}. In Conditional Random Fields (CRFs), feature functions act like preference rules. In addition, priorities (weights) for each feature function are automatically learned with the labeled data. 

Furthermore, Chen and Su developed a harmonic analyzer called the Harmony Transformer that used a Transformer encoder-decoder as an input feature extractor\cite{chen_ht,chen_att}. More recently, Micchi et al. proposed a model with a Convolutional Recurrent Neural Network (CRNN) encoder and Neural Autoregressive Distribution Estimator (NADE) that outperformed the Harmony Transformer\cite{micchi_rome,micchi_deep}. However, even with recent neural network-based supervised learning, the performance of harmonic analysis still has room to be improved. In addition, the dataset is biased toward piano pieces in the classical era and vocal pieces with relatively clear harmony.

\subsubsection{Unsupervised harmonic analysis}
Increasing the amount of training data is a way to improve performance in supervised learning. However, preparing the data can be expensive. Alternatively, unsupervised learning does not require labeled training data. However, there have been few works on unsupervised harmonic analysis\cite{raphael_gra,wang2012}. The challenge with harmonic analysis is that it involves simultaneously identifying both keys (determined by the combination of modes and tonics) and chords. Since many possible combinations exist, obtaining an optimal result through unsupervised learning is challenging. 
Consequently, past studies had to rely on manually set model parameters\cite{raphael_gra,wang2012}.
For instance, Wang and Wechsler proposed using an Infinite Gaussian Mixture Model (IGMM) for harmonic analysis\cite{wang2012}. In this model, keys and chords were considered as hidden variables. Since IGMM is a type of Bayesian model, optimal keys and chords for targeted musical notes are obtained through the sampling process. One limitation of their model was that the model parameters were given manually based on musical knowledge. For example, they provided the Gaussian mean and covariance for the major and minor scales, which worked like key profiles. In addition, the IGMM could not consider chord transitions since it was a note clustering.

Raphael and Stoddard also proposed an unsupervised harmonic analysis approach based on Hidden Markov Models (HMMs)\cite{raphael_gra}. The HMM had the advantage of being able to account for chord transitions. However, according to Raphael and Stoddard, while the chord degree transition probability was learnable, the key transition probability and the probability of the chord degree after a modulation were difficult to learn\cite{raphael_gra}. This difficulty is probable because the chord degrees appear relatively evenly, whereas keys and modulations strongly depend on individual pieces.

\subsection{Deep latent variable models}
Deep latent variable models are methods in which deep neural networks work as the approximators of the probability distributions that make up the latent variable models. \cite{kim2018,tran2016,miao2017}. In recent years, deep latent variable models have attracted attention as a new method for improving unsupervised learning. For example, Tran et al. proposed unsupervised neural Hidden Markov Models for the part-of-speech tag induction\cite{tran2016}, and Miao et al. introduced neural topic models\cite{miao2017}. 

In the music field, Uehara and Tojo extended the unsupervised neural HMMs to semi-Markov models (HSMMs) and performed recognition of chord segments and their transitions\cite{uehara2022}. However, their model was not able to recognize keys and chords simultaneously, and thus could not perform harmonic analysis. Our model is based on Uehara's Neural HSMMs but with significant changes for harmonic analysis, including output emission probability with chord quality templates, simultaneous recognition of keys and roots, and dynamic modulation detection.

\section{Methodology}
\label{sec:methodology}
The proposed model for harmonic analysis requires no labeled data for parameter training. In this sense, we describe the model as "unsupervised," which is a significant step forward, given that conventional models require all or part of the model parameters to be designed manually\cite{raphael_gra,wang2012}. However, our model incorporates simplifications proposed in previous methods, such as the assumption of transpositional equivalency, a predefined set of chord qualities, and simplified conditional probabilities of the stochastic model.

The hidden semi-Markov model\cite{yu2010} forms the core of the proposed model. Unlike Markov models, semi-Markov models equip a notion of the duration of a state, thus more suitable for segment-level recognition. This property of the HSMM is advantageous since chords are recognized as a result of score segmentation\cite{masada_crf}. Among several variants of the hidden semi-Markov model, we utilize the "Residential-time HMM"\cite{yu2003} that assumes independence between the duration of the current and the previous hidden states.

The EM algorithm is a widely known method for learning the parameters of an HSMM\cite{yu2003,yu2010}. However, it has been reported that Neural HSMM, a type of deep latent variable model, could achieve better marginal likelihood than the EM algorithm\cite{ueharathesis}. As described in Section 2, the deep latent variable model predicts model parameters through neural networks\cite{kim2018}. In the following sections, we will first overview the proposed HSMM for harmonic analysis. Then, the neural networks that approximate probability distributions that constitute HSMM are described.
Table~\ref{tab:notation} shows the notations that will be used in the rest of this paper.
\begin{table}[htb!]
\centering
\small
\caption{Notations.}
    {\begin{tabular}{c|l|l}
    \hline
    $L$ & Output sequence length & $\in \mathbb{N}$ \\
    $t$ & Time step & $\in \{0, \dots, L-1\}$ \\
    $\bm{x}_t$ & An observation at $t$ & $\in \{0, 1\}^{12}$ \\ 
    $pc$ & Pitch class\footnote{The pitch classes are the 12 numbered notes in an octave: \{C, C\#/Db, D, ..., B\} are numbered \{0, 1, 2, ..., 11\} respectively.} & $\in \{0, \dots, 11\}$ \\
    $m$ & Mode index & $\in \{0, 1\}$ \\
    $r$ & Root pitch class index\footnote{\label{foot:root}In our design, the root pitch class represented by the index $r$ indicates the root note of the chord, independent of the key.}\footnote{The last 13th dimension is used as the \textit{Rest} state.} & $\in \{0, \dots, 12\}$ \\
    $i, j$ & \scriptsize{Root pitch class indices that distinguish before/after transition.} & $\in \{0, \dots, 12\}$\\
    $s$ & Shift value. ($m$, $s$) specifies a key $k$. & $\in \{0, \dots, 11\}$ \\
    $k$ & Key index & $\in \{0, \dots, 23\}$ \\
    $q$ & Chord quality index & $\in \{0, \dots, 6\}$ \\
    $d$ & Hidden state duration index & $\in \{0, \dots, 15\}$ \\
    $N_{(\cdot)}$ & The number of indices $(\cdot)$ & $\in \mathbb{N}$ \\
    $v_{(\cdot)}$ & \textit{Logit}\footnote{In this paper, we use the term \textit{logit} as a value that is derived from a neural network and parameterize a probability. The logit is fed into the sigmoid or softmax function to produce the parameter of a targeted probability distribution.} for the probability of $(\cdot)$ & $\in \mathbb{R}$ \\
    $\bm{v}_{(\cdot)}$ & Vector of logits for the probability of $(\cdot)$ & $\in \mathbb{R}^{N_{(\cdot)} - 1}$ \\
    $\bm{z}_{(\cdot)}$ & Index vector with the \textit{1 of} $N$ \textit{representation}\footnote{The \textit{1 of} $N$ \textit{representation} is an $N$-dimensional vector representation of a category index. For example, for chord quality index $q=3$, the corresponding vector of \textit{1 of} $N$ \textit{representation} is $\{0, 0, 0, 1, 0, 0, 0\}$.} & $\in \{0, 1\}^{N_{(\cdot)}}$ \\
    $z_{(\cdot)}$ & The $(\cdot)$-th component of the 1 of $N$ index vector. & $\in \{0, 1\}$\\
    $\bm{e}_m$ & Mode embedding & $\in \mathbb{R}^{12}$ \\
    \hline
    \end{tabular}}
\label{tab:notation}
\end{table}

\subsection{The hidden semi-Markov model for harmonic analysis}
The proposed model is based on a hidden semi-Markov model comprising initial hidden-state, hidden-state transition,  hidden-state duration, and output emission distributions\cite{yu2003,yu2010,uehara2022}. The output for the model is a sequence of pitch classes represented in binary twelve-dimension vectors. Since the hidden states are not observed and no labeled data is used, what the hidden states represent is not specified. However, simplifications of the model, described later in this paper, make the hidden states correspond to combinations of keys and root pitch classes.

\begin{figure}[hbt!]
\centering
\includegraphics[width=\linewidth]{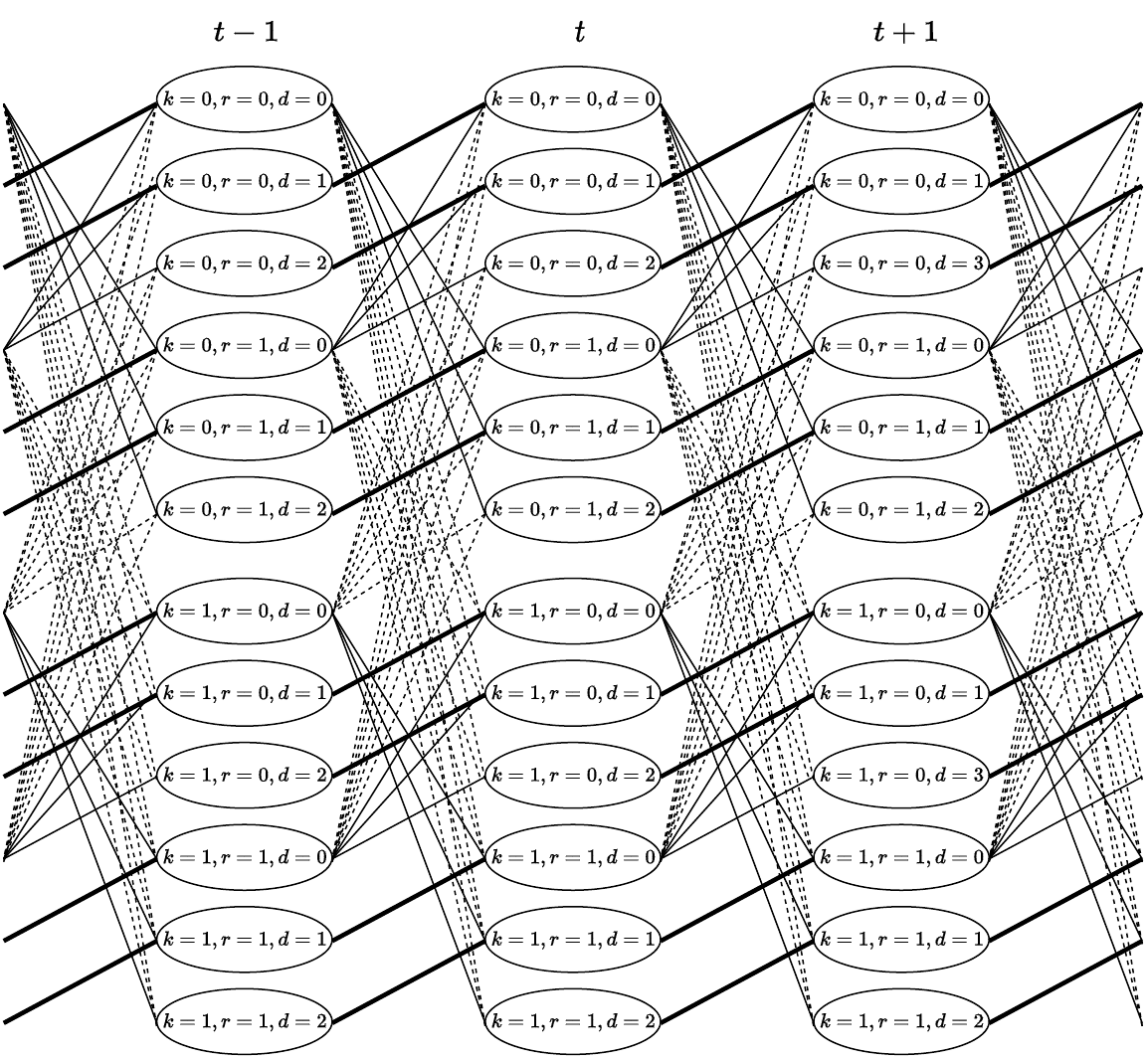}
\caption{An example of possible paths of hidden states where the number of keys is 2, the number of roots is 2, and the maximum duration is 3. For simplicity, the number of keys, roots, and the maximum duration are less than the actual settings. Thin solid lines represent root transitions, dotted lines represent key transitions (modulations), and bold solid lines represent the consumption of the remaining duration times.}
\label{fig:possible-state-path}
\end{figure}
Figure~\ref{fig:possible-state-path} shows an example of possible paths of hidden states. Note that the numbers of keys, root pitch classes, and the maximum duration are less than the actual settings for ease of reading the figure. At each time step, the hidden states are represented by a combination of a key and a root pitch class. In addition, possible remaining times of the hidden states are combined with each hidden state. Then, triplets of (key, root, duration) represent all possible states. Key or root transitions are only permitted when the remaining duration time is zero in a semi-Markov model. Otherwise, the remaining duration times decrease by one at each time step. Note that the Residential-time HMM, a variant of HSMMs, assumes a hidden state transition is independent of the duration (residential-time) of the previous hidden state\cite{yu2003,yu2010}. Thus, the hidden state transition is represented as follows. 
\begin{equation}
\small
    \label{eq:hsmm-state-transition}
    p (k_t, r_t, d_t | k_{t-1}, r_{t-1}, d_{t-1}) =
    \begin{cases}
        p(k_t,r_t,d_t | k_{t-1}, r_{t-1})~~(\text{if~} d_{t-1} = 0) \\
        {\mathds{1} (d_t = d_{t-1} - 1)~~(\text{if~} d_{t-1} > 0)}
    \end{cases}
\end{equation}
Note that transition probabilities do not include self-transitions, as illustrated in Figure~\ref{fig:possible-state-path}.
In addition, we simplify the hidden-state transition probability $p(k_t,r_t,d_t | k_{t-1}, r_{t-1})
$ as follows.
\begin{equation}
    \label{eq:decomposed-state-transition}
    p(k_t,r_t,d_t | k_{t-1}, r_{t-1}) = 
    \begin{cases}
        {p(d_t) p(r_t | k_t) p(k_t|k_{t-1})~~(\text{if~} k_t \neq k_{t-1})} \\
        {p(d_t) p(r_t | r_{t-1}, k_t) p(k_t | k_{t-1})~~(\text{if~} k_t = k_{t-1})}
    \end{cases}
\end{equation}
In the rest of the paper, we call $p(k_t|k_{t-1})$ as the key transition probability, $p(r_t | k_t)$ as the initial root probability, $p(r_t | r_{t-1}, k_t)$ as the root transition probability, and $p(d_t)$ as the duration probability.

Since there is no previous hidden state when $t=0$, the probability of the first hidden state is computed as follows.
\begin{equation}
p(k_0, r_0, d_0) = p(d_0) p(r_0 | k_0) p(k_0)
\label{eq:initial-prob}
\end{equation}
The distribution for $p(d_0)$ is the same as $p(d_t)$, and $p(r_0 | k_0)$ is the same as $p(r_t | k_t)$. $p(k_0)$ is the initial key probability.

At each time step, the output $\bm{x}_t$ depends only on the hidden state at the same time step. We decompose the emission probability $p(\bm{x}_t | k_t, r_t)$ into a chord quality probability $p(q_t | k_t, r_t)$ and the pitch class emission probability $p(\bm{x}_t | q_t, r_t)$, where $q_t$ denotes a chord quality.
\begin{equation}
p(\bm{x}_t | k_t, r_t) = \sum_{q_t} p(\bm{x}_t | q_t, r_t) p(q_t | k_t, r_t)
\label{eq:emission-probability}
\end{equation}
Among all the probability distributions in the proposed HSMM, we only set manually $p(\bm{x}_t | q_t, r_t)$, which we call \textit{chord quality templates}\footnote{More precisely, in addition to this, the maximum inter-key transition probability limit is set manually.}. The details of the chord quality templates are described in \ref{sec:emission-distribution}.

With the initial(\ref{eq:initial-prob}) and transition(\ref{eq:hsmm-state-transition})(\ref{eq:decomposed-state-transition}) probability of the hidden states and the emission probability of the observations(\ref{eq:emission-probability}) described above, the generative process of the HSMM is represented as follows, where $L$ is the sequence length.
\begin{align}
& p(\bm{x}_0, \dots, \bm{x}_{L-1}, k_0, \dots, k_{L-1}, r_0, \dots, r_{L-1}, d_0, \dots d_{L-1})  \nonumber\\
&= P(k_0, r_0, d_0) \prod^{L-1}_{t=0} P(\bm{x}_t | k_t, r_t) \prod^{L-1}_{t=1} P(k_t, r_t, d_t | k_{t-1}, r_{t-1}, d_{t-1})
\end{align}

\subsubsection{Emission distribution with chord quality templates}
\label{sec:emission-distribution}
In this section, we first describe the details of the calculation of output emission distribution. As described above, we decompose the output emission probability into the chord quality and pitch class emission probability. We use seven chord qualities: \{major triad, minor triad, diminish triad, dominant seventh, major seventh, minor seventh, and diminish seventh\}. Then, we formulate the probability of pitch class emission given the chord quality and the root as follows.
\begin{align}
p(\bm{x}_t|q_t, r_t) &= \prod^{11}_{pc=0} \textrm{Bernoulli}(x_{t,pc} | q_t, r_t) \nonumber\\
&= \prod^{11}_{pc=0} (\mu_{pc | q, r})^{x_{t,pc}} (1 - \mu_{pc | q, r})^{1 - x_{t,pc}} \\
\mu_{pc | q, r} &= \textrm{sigmoid}(v_{pc | q, r})
\label{eq:mu}
\end{align}
$x_{t,pc}$ represents the value at the pitch class ($pc$) of an output binary vector $\bm{x}_t$.
For example, if $\bm{x}_t = \{0, 0, 1, 0, 0, 0, 1, 0, 0, 1, 0, 0\}$, the values of $x_{t,2}$, $x_{t,6}$, and $x_{t,9}$ are 1, and the remaining are 0.

\begin{table}[hbt]
\centering
\small
\caption{Settings of chord quality templates: $v_{pc | q,0}$ (root pithch class = 0). Where one(1) stands, $v_{pc | q,0} = w$, and $v_{pc | q,0} = -w$ otherwise.}
\label{tab:chord-quality-templates}
\begin{tabular}{l|cccccccccccc}
\hline
quality ($q$) & pc=0 & 1 & 2 & 3 & 4 & 5 & 6 & 7 & 8 & 9 & 10 & 11 \\ \hline
major triad ($\textrm{M}$) & 1 &  &  &  & 1 &  &  & 1 &  &  &  & \\ \hline
minor triad ($\textrm{m}$) & 1 &  &  & 1 &  &  &  & 1 &  &  &  & \\ \hline
diminish triad ($\textrm{d}$) & 1 &  &  & 1 &  &  & 1 &  &  &  &  & \\ \hline
dominant ($7$) & 1 &  &  &  & 1 &  &  & 1 &  &  & 1 & \\ \hline
major seventh ($\textrm{M}_7$) & 1 &  &  &  & 1 &  &  & 1 &  &  &  & 1\\ \hline
minor seventh ($\textrm{m}_7$) & 1 &  &  & 1 &  &  &  & 1 &  &  & 1 & \\ \hline
diminish seventh ($\textrm{d}_7$) & 1 &  &  & 1 &  &  & 1 &  &  & 1 &  & \\ \hline
\end{tabular}
\end{table}
We set the logit $v_{pc | q,r}$ manually, which we call \textit{chord quality templates}, equivalent to setting chord labels in supervised learning.
In particular, we set the quality template for the case root pitch class of 0, as in Table~\ref{tab:chord-quality-templates}.
$v_{pc | q,0} = w$ when marked by one(1) in Table~\ref{tab:chord-quality-templates}, and $v_{pc | q,0} = -w$ otherwise. The weight $w$ is set to $5.0$.
As described in Table~\ref{tab:notation}, we use a bold symbol to represent a vector of logits, for example, 
\begin{equation}
\bm{v}_{pc|q,r} = (v_{pc=0|q,r}, v_{pc=1|q,r}, \dots, v_{pc=11 | q,r})^\top \in \mathbb{R}^{12}.
\end{equation}

If the root pitch class is not 0, the chord templates for a specific root pitch class are obtained simply by shifting the code templates in Table~\ref{tab:chord-quality-templates}.
For example, when the root pitch class is 5, the chord template of minor seventh $\bm{v}_{pc|q=\textrm{m}_7,r=5} = \{w, -w, -w, w, -w, w, -w, -w, w, -w, -w, -w\}$, where $v_{pc=0 | q=\textrm{m}_7, r=5}$, $v_{pc=3 | q=\textrm{m}_7, r=5}$, $v_{pc=5, | q=\textrm{m}_7, r=5}$, and $v_{pc=8 | q=\textrm{m}_7, r=5}$ are $w$ and the remaining are $-w$.

On the other hand, the probability of chord quality given a key and a root pitch class is calculated via neural networks as follows\footnote{${\rm softmax}_j(\bm{x}) = \frac{\exp({x_j})}{\sum_{j'} \exp({x_j'})}$}.
\begin{align}
p(\bm{z}_{q|m,r} | m, r) &= \textrm{Categorical}(\bm{z}_{q|m,r} | m, r) = \prod^{N_q - 1}_{q=0} \pi^{z_{q|m,r}}_{q|m,r} \nonumber\label{eq:chord-quality-prob}\\
\pi_{q|m,r} &= p(q|m,r) = \textrm{softmax}_q(V^\top_{pc,q|r} \bm{e}_{m,r}) \\
\bm{e}_{m, r} &= (\bm{e}_{m \times r})_r \in \mathbb{R}^{12} \label{eq:select-expand-mode-embedding}\\
\bm{e}_{m \times r} &= W^\top_{m \rightarrow m \times r} \bm{e}_m \in \mathbb{R}^{12 \times 12} \label{eq:expand-mode-embedding}
\end{align}
Here, $V_{pc,q|r} \in \mathbb{R}^{12 \times N_q}$ is a set of chord templates $\bm{v}_{pc|q,r} \in \mathbb{R}^{12}$, represented with a tensor.
Equation (\ref{eq:expand-mode-embedding}) represents the operation of expanding the dimension of mode embedding by a trainable linear transformation to the number of modes times the number of roots, then, eq. (\ref{eq:select-expand-mode-embedding}) represents extracting the components corresponding to the targeted root.
The computation of the mode embeddings $\bm{e}_m$ will be explained in the following paragraph.

Since the probability $p(q | m, r)$ is independent on the time step $t$, the subscript $t$ is omitted in (\ref{eq:chord-quality-prob}). Hereafter, we sometimes omit the subscript $t$ when the variable is time-independent. Note that the chord quality probability is conditioned on the mode $m$ and root $r$ instead of key $k$ and root $r$. In this paper, we set the number of modes to two. However, the two modes are not restricted to major and minor modes but are automatically learned by unsupervised learning. 

\begin{align}
\bm{h}_m &= \textrm{LSTM}(\bm{e}_{m-1}, \bm{h}_{m-1}) \nonumber\\
\bm{e}_m &= \bm{h}_m
\label{eq:mode-embedding}
\end{align}
The $\bm{e}_m \in \mathbb{R}^{12}$ in equation (\ref{eq:expand-mode-embedding}) is the mode embedding, which works as a latent feature vector of a mode. 
This $\bm{e}_m$ can be a learnable vector but is generated through a Recurrent Neural Network (RNN) in this study, as shown in (\ref{eq:mode-embedding}). In particular, we utilize the Long-Short Term Memory\cite{hochreiter1997} as an RNN. In this way, any number of mode embeddings can be generated with the same RNN. In this study, the number of modes is fixed to two, so there is no direct advantage to using an RNN, but this setting is advantageous if the number of modes is larger in the future (e.g., considering church modes) or if it is difficult to fix the number of modes in advance.

Once $p(q|m,r)$ is calculated, $p(q|k,r)$ is obtained by shifting $p(q|m,r)$, where we assume transpositional equivalence in keys.
For example, a key with (mode index = 1,  shift = 3) is assumed to represent a key, each pitch class of which is shifted by three from (mode index = 1, shift = 0).
We let the symbol $m$ denote the shift($s$) = 0 case, and let $k$ denote $k=0: (m=0, s=0)$, $k=1: (m=0, s=1)$, ... , $k=11: (m=0, s=11)$, $k=12: (m=1, s=0)$, $k=13: (m=1, s=1)$, ... , $k=23: (m=1, s=11)$.
Note that we do not restrict the tonic pitch class of a mode as $r=0$.
We will describe how we get the tonic pitch class later.
The shift of $p(q|m,r)$ is done without the tonic information.
For example, when $k=14$, $p(q|k=14,r=2) = p(q|m=1,r=0)$.

As we have seen, the parameters in equations (\ref{eq:expand-mode-embedding}) and (\ref{eq:mode-embedding}) are learnable and are utilized to generate the chord quality distribution. Thus, the neural networks generate the distributions that make up the model, which is a characteristic feature of the deep latent variable model\cite{kim2018}. The mode embedding $\bm{e}_m$ is also used to calculate the initial hidden state and the hidden state transition probabilities described below. The results of previous studies suggest that using such features and network elaborations can lead to better convergence than conventional methods\cite{tran2016,miao2017,uehara2022}.

\subsubsection{Hidden state transition distribution}
A hidden state is represented in a combination of key, root pitch class, and remaining duration. As shown in equations (\ref{eq:hsmm-state-transition}) and (\ref{eq:decomposed-state-transition}), the hidden state transition probability is decomposed in the duration probability $p(d_t)$, root transition probability $p(r_t|r_{t-1},k_t)$, and key transition probability $p(k_t|k_{t-1})$.

In this work, we assume $p(d)$ is independent of keys and roots since coherent chord length would be preferred.
Therefore, the logit for $p(d)$ is a simple learnable vector, the dimension size of which is the maximum duration length.
Thus, the duration distribution is obtained as follows, where $\bm{v}_d \in \mathbb{R}^{N_d}$ is the learnable vector.
\begin{align}
\label{eq:duration-prob}
p(\bm{z}_d) &= \textrm{Cat.}(\bm{z}_d) = \prod^{N_d - 1}_{d=0} \pi^{z_d}_d \\ 
\pi^{z_d}_d &= p(d) = \textrm{softmax}_d(\bm{v}_d)
\end{align}

For the root transition probability $p(r_t=j|r_{t-1}=i, k)$, we first compute $p(r_t=j|r_{t-1}=i, m)$ as follows, then shift them under the assumption of transpositional equivalency.
\begin{align}
p(\bm{z}_{j|i,m}|r_{t-1}=i, m) &= \textrm{Cat.}(\bm{z}_
{j|i,m}|r_{t-1}=i, m) = \prod^{12}_{j=0} \pi^{z_{j|i,m}}_{j|i,m} \\
\pi^{z_{j|i,m}}_{j|i,m} &= p(r_t=j|r_{t-1}=i, m) = \textrm{softmax}_{j}(\bm{v}_{j|i,m}) \label{eq:root-transition-prob}\\
v_{r_t=j|r_{t-1}=i, m} &= \textrm{MLP}_2(\bm{e}_m \oplus \bm{v}_{pc|i,m} \oplus \bm{v}_{pc|j,m}) \label{eq:root-transition-logit} \\
\bm{v}_{pc|r,m} &= \sum_q p(q|m, r) \bm{v}_{pc|q,r} \label{eq:marginal-emission-logit}
\end{align}
The equation (\ref{eq:marginal-emission-logit}) computes a marginal emission pitch class logit of the root pitch class $r$. The marginal emission pitch class logit $\bm{v}_{pc|r,m} \in \mathbb{R}^{12}$ is used as a feature of a chord whose root is $r$ given the mode $m$. We consider the last root index ($r=12$) as \textit{Rest} and set $\bm{v}_{pc|r,m}$ for it to a 12-dimensional vector with each dimension's value as $-w$. Then, equation (\ref{eq:root-transition-logit}) computes a root-transition logit $v_{r_t=j|r_{t-1}=i, m}$, where $\oplus$ denotes a vector concatenation.
The root transition probability $p(j|i,m)$ is then computed by the softmax on the set of logits $\bm{v}_{j|i,m} = (v_{j=0|i,m}, \dots, v_{j=12|i,m})$.

Once $p(r_t|r_{t-1}, m)$ is computed, $p(r_t|r_{t-1}, k)$ is obtained by shifting $p(r_t|r_{t-1}, m)$, similar to the above discussion of the chord quality probability case.
For example, when $k=14$, which means $(m=1, s=2)$, a component of the probability $p(r_t=9|r_{t-1}=2, k=14)$ equals $p(r_t=7|r_{t-1}=0, m=1)$.

We use the same emission distribution, duration distribution, and root transition distribution for all observed data. On the other hand, the key distributions described hereafter can take different values for each observation sequence. Changing the probability distribution of the model depending on the observation is not possible with a conventional HSMM, but is possible with deep latent variable models\cite{tran2016,kim2018,miao2017,uehara2022}.

We first obtain the embedding of an observed sequence to compute the key transition distribution.
\begin{align}
{\bm h}^\textrm{obs.}_t &= {\rm LSTM}^\textrm{obs.} \bigr( {\rm tanh}(W^\top {\bm x}_t), {\bm h}^\textrm{obs.}_{t-1} \bigr) \label{eq:obs-enc-lstm}\\
\hat{a}_t &= \textrm{MLP}_2({\bm h}^\textrm{obs.}_t \oplus t^{ratio}) \label{eq:obs-enc-attn1}\\
\alpha_t &= \frac{\exp{(\hat{a}_t})}{\sum^{L-1}_{t'} \exp{(\hat{a}_{t'})}} \label{eq:obs-enc-attn2}\\
\hat{{\bm h}}^\textrm{obs.} &= \sum_t \alpha_t {\bm h}^\textrm{obs.}_t \label{eq:obs-enc}
\end{align}
The observed data at each time step $\bm{x}_t$ is first embedded by an LSTM, as eq. (\ref{eq:obs-enc-lstm}). Next, the weight for each latent feature $\bm{h}^\textrm{obs.}_t \in \mathbb{R}^{12}$ is calculated by (\ref{eq:obs-enc-attn1}) and (\ref{eq:obs-enc-attn2}), where the scalar value $t^{ratio} = \frac{t}{L}$ is an additional information of the time step. Then, the embedding of the observed sequence is obtained by the weighted summation of the latent features, as shown in (\ref{eq:obs-enc}).
Using the resulting $\hat{\bm{h}}^\textrm{obs.} \in \mathbb{R}^{12}$, we compute the mode probability as follows.
\begin{align}
p(\bm{z}_{m}) &= \textrm{Cat.}(\bm{z}_m) = \prod^{N_m - 1}_{m=0} \pi^{z_m}_m \\
\pi^{z_m}_m &= p(m) = \textrm{softmax}_m(E^\top \hat{\bm{h}}^\textrm{obs.}) \label{eq:mode-prob}
\end{align}
Here, $E \in \mathbb{R}^{12 \times N_m}$ is a set of mode embeddings $\bm{e}_m$, represented with a tensor.

As described in Section~\ref{sec:emission-distribution}, a key is defined by specifying a combination of mode index $m$ and shift value $s$. The distribution of modes is obtained by (\ref{eq:mode-prob}), and the shift value is obtained as follows.
\begin{align}
p(\bm{z}_{s|m}|m) &= \textrm{Cat.}(\bm{z}_{s|m}|m) = \prod^{11}_{s=0} \pi^{z_{s|m}}_{s|m} \\
\pi^{z_{s|m}}_{s|m} &= p(s|m) = \textrm{softmax}_s(\textrm{MLP}_2(\bm{e}_m \oplus \hat{\bm{h}}^\textrm{obs.})) \label{eq:shift-prob}
\end{align}
Then, the following equation gives the key probability.
\begin{align}
p(k) &= p(m, s) = p(s|m) p(m) \label{eq:initial-key-prob}
\end{align}
Note that the key probability $p(k)$ is not conditioned on the previous key; therefore, it is used as the initial key probability.
There are two situations where the initial key probability is used, one at the beginning of the sequence and the other immediately after the modulation.
Although the transition probabilities between keys should be considered, the number of parameters to be estimated was reduced by computing $p(k)$ instead of $p(k_t|k_{t-1})$, as the proportions are more important than the transitions for the keys. The key transition probability is simplified as follows.
\begin{equation}
    \label{eq:key-transition}
    p (k_t | k_{t-1}) =
    \begin{cases}
        {1 - \beta~~(\text{if~} k_{t} = k_{t-1})} \\
        {\beta~\hat{p}(k)~~(\text{if~} k_t \neq k_{t-1})}
    \end{cases}
\end{equation}
$\beta$ is a learnable value. However, we set the upper limit of $\beta$ to $0.01$, which means the model penalizes modulations. $\hat{p}(k)$ is the key probability modified so that $k_t \neq k_{t-1}$.

At the beginning of the sequence, the key probability $p(k)$ is combined with the initial root probability given a key in the following equation.
Here, again, we first compute the initial root probability given a mode $p(r|m)$, and the initial root probability given a key is obtained automatically by shifting $p(r|m)$.
\begin{align}
p(\bm{z}_{r|m}|m) &= \textrm{Cat.}(\bm{z}_{r|m}|m) = \prod^{12}_{r=0} \pi^{z_{r|m}}_{r|m} \\
\pi^{z_{r|m}}_{r|m} &= p(r|m) = \textrm{softmax}_r (\bm{v}_{r|m}) \\
v_{r|m} &= \textrm{MLP}_2(\bm{e}_m \oplus \bm{v}_{pc|r,m})
\end{align}
In the above equation, $\bm{e}_m$ is the mode embedding, and $\bm{v}_{pc|r,m}$ is the marginal emission pitch class logit, the same as those used in (\ref{eq:marginal-emission-logit}) and (\ref{eq:root-transition-logit}).
The Multi-Layer Perceptron (MLP) with one hidden layer maps the concatenated vector of $\bm{e}_m \oplus \bm{v}_{pc|r,m}$ to a scalar logit $v_{r|m}$.
The initial probability given a mode $p(r|m)$ is then computed by applying the softmax function to the set of logits $\bm{v}_{r|m} = (v_{r=0|m}, \dots, v_{r=12|m})$.
The initial root probability given a key $p(r|k)$ can be automatically computed by shifting the $p(r|m)$.
Finally, we can obtain the initial probability by $p(k, r, d) = p(d) p(r|k) p(k)$.

\subsection{Training}
\label{sec:training}
As a deep latent variable model, the proposed model is trained by optimizing the network parameters that produce the probability distributions of the HSMM.
The loss function here is the negative log-likelihood (NLL) of the observed sequence $-\log p(\bm{x}_0, \cdots, \bm{x}_{L-1})$\footnote{More precisely, the loss is the average of the NLL of all the sequences in the mini-batch.}, obtained by marginalizing all possible paths of hidden states.
The technique to marginalize all possible paths of hidden states for an H(S)MM is known as the \textit{forward algorithm}\cite{rabiner1989, yu2010}.
The details of the forward algorithm for the HSMM, especially for the Residential-time HMM, can be found in \cite{yu2003} and \cite{uehara2022}.
In the proposed model, a hidden state is represented in a combination of key and root, as illustrated in Figure~\ref{fig:possible-state-path}. Then, the model can directly apply the forward algorithm for the Residential-time HMM, described in \cite{uehara2022}.

We perform unsupervised training in two phases.
The proposed model is not given a feature of keys like the \textit{Key Profile}\cite{krumhansl2001cognitive}. Therefore, in the first phase, we train the score without key signatures to obtain features of two modes. After that, additional training is performed using the original key signature.
\begin{itemize}
\setlength{\leftskip}{8mm}
\item[Phase 1:] Training with the \textit{normalized} data. \textit{Normalized} means transposing a score to a key without a key signature. If the key signature information has been lost, the score is transposed to maximize the percentage of pitch classes \{0, 2, 4, 5, 7, 9, 11\}. In Phase 1 training, keys are limited to the two modes only. These two modes are not necessarily limited to C major and A minor keys, but are learned under the condition that the shift value is always 0.
In addition, modulation is disabled by setting the maximum inter-key transition probability limit to zero.
\item[Phase 2:] Training with the original (not transposed) data. In the Phase 2 training, the shift values and the inter-key transition are enabled. Phase 2 training is performed as additional training using the result of Phase 1 as the initial value.
\end{itemize}

\subsection{Inference (harmonic analysis)}
After the training, the proposed model performs harmonic analysis by finding the most likely sequence of hidden states (combinations of keys and roots) and their residential times from an observed sequence.
Thus, the harmonic analysis is the inference problem of an HSMM, and the method for it is well known as the Viterbi algorithm\cite{forney1973,rabiner1989,yu2010}.

After the hidden state inference, the most likely chord quality ($q_t^*$) for each timestep is obtained by taking the argmax of the probability of output emission given a key and a root.
\begin{align}
q_t^* &= \arg \max_q p(\bm{x}_t|q, r_t)p(q|k_t, r_t)
\end{align}
Thus, we obtain the sequence of keys, root pitch classes, and qualities.
Both chord name analysis and Roman numeral analysis are possible based on them.
In particular, a chord name is derived from the information of a root pitch class\footnote{$\rightarrow$ footnote~\ref{foot:root}} and a quality.
The Roman numeral is derived from a key, root, and quality.

To obtain the chord degree (Roman numeral), we first convert the root pitch class by $(r - s) \mod 12$. Here, $s$ is the shift value that is combined with mode $m$ to specify a key ($k = (m, s)$).
Although our model does not explicitly know the diatonic scale, the transition probabilities of the learned model indicate that the chords with 0 and 7 as the roots are the most dominant. 
This result will be discussed later, but based on this observation, the correspondence between pitch class and degree is given in Table~\ref{tab:convert-to-degree}. Here, the pitch classes corresponding to I and V are 0 and 7, respectively, and the rest are assigned one degree per two successive roots. After the conversion, the degree is denoted in uppercase if the predicted quality is major; otherwise, it is denoted in lowercase.
\begin{table}[hbtp]
\centering
\small
\caption{Conversion table of shifted root pitch classes to degrees.}
\label{tab:convert-to-degree}
\begin{tabular}{l|cccccccccccc}
\hline
Shifted root pc. & 0 & 1 & 2 & 3 & 4 & 5 & 6 & 7 & 8 & 9 & 10 & 11 \\ \hline
Degree & I & II & II & III & III & IV & IV & V & VI & VI & VII & VII \\ \hline
\end{tabular}
\end{table}

Our model does not predict the chord inversion. However, we determine chord inversions in Table~\ref{tab:inversion}, where the blank components are assumed to be the basic chords.
\begin{table}[htbp]
\centering
\scriptsize
\caption{Conversion table for determining chord inversions.}
\label{tab:inversion}
\begin{tabular}{l|cccccccccccc}
\hline
(bass pc. - $r$) $\mod$ 12 & 0 & 1 & 2 & 3 & 4 & 5 & 6 & 7 & 8 & 9 & 10 & 11 \\ \hline
major triad ($\textrm{M}$) & & & & & & 6/4 & & & 6 & & & \\ 
minor triad ($\textrm{m}$) & & & & & & 6/4 & & & & 6 & & \\
diminish triad ($\textrm{d}$) & & & & & & & 6/4 & & & 6 & & \\
dominant ($7$) & & & 2 & & & 4/3 & & & 6/5 & & & \\
major seventh ($\textrm{M}_7$) & & 2 & & & & 4/3 & & & 6/5 & & & \\
minor seventh ($\textrm{m}_7$) & & & 2 & & & 4/3 & & & & 6/5 & & \\
diminish seventh ($\textrm{d}_7$) & & & & 2 & & & 4/3 & & & 6/5 & & \\
\hline
\end{tabular}
\end{table}

\section{Experiments}
\label{sec:experiments}
\subsection{Datasets}
We used two different sets of J.S. Bach's four-part chorales: a dataset of 60 chorales formatted by Radicioni and Esposito, and a set of 371 chorales in MusicXML format from the Music21 Library. In this paper, we denote the former dataset as "JSBChorales60" and the latter as "JSBChorales371".

Radicioni and Esposito provided preprocessed scores and human-annotated chord labels in the JSBChorales60 dataset\footnote{\url{https://archive.ics.uci.edu/dataset/298/bach+choral+harmony}}\cite{radicioni2010}. The preprocessed scores include pitch classes, bass pitch classes, and metrical accents computed by the Meter program of the Melisma Analyzer\cite{temperley_web}. However, the preprocessing lost some information in the original scores, for example, beat positions, key signatures, time signatures, and parts (except bass pitch classes). The dataset contains 60 chorales in total.

On the other hand, The JSBChorales371 dataset from the Music21 Library\cite{music21} is a set of scores in MusicXML format, and it retains all information as a score. In addition, the dataset contains 371 chorales, some of which are not included in the JSBChorales60 dataset. One of the reasons the JSBChorales371 dataset has not been used much is that it is just a collection of scores and does not provide labeled data. Since our model is unsupervised, we can utilize the JSBChorales371 dataset. For evaluation, we consult the human analysis of 20 pieces publicly available in the Music21 Library\footnote{\label{footnote:chorale-analysis}\url{https://github.com/cuthbertLab/music21/tree/master/music21/corpus/bach/choraleAnalyses}}\cite{music21}. We used fixed train, validation, and test splits for the JSBChorales371 dataset. The testing data is the set of 20 pieces where the human annotations are available. We preprocess the MusicXML scores into sequences of pitch classes. The length of each event of pitch classes is 16th-note width, which is the minimum duration of the dataset (except for very few exceptions). 

\subsection{Experimental setups}
\label{sec:setups}
For the JSBChorales60 dataset, we followed the original 10-fold cross-validation splits provided by Masada and Bunescu. However, unlike the previous work, we used one fold for testing, another for development, and the remaining 8-folds for training. The random seed was fixed to 123. To train the proposed model, we only used the preprocessed pitch classes. The minibatch size was 2.
Since only the annotation of chord names is provided, we performed chord name analysis for the JSBChorales60 dataset. In addition, the chord root names in the annotation were converted to pitch classes before the evaluation since our model did not distinguish enharmonic notes. However, we do not consider this a serious limitation since enharmonic distinctions are possible if the original score information has not been lost in preprocessing.

For the JSBChorales371 dataset, we used fixed train, validation, and test splits for the JSBChorales dataset. The testing data was the set of 20 pieces where the human annotations were available. However, three pieces were excluded because of the collapsed format or inconsistency of key signatures between the annotation and the original score. Thus, the resulting number of testing data was 17. In addition, we used 62 pieces randomly selected as the development data and the remaining 243 pieces for training. The total number of pieces used was 322, where 49 pieces were excluded because of the collapsed format or duplication. In both training and testing, we separated a piece into segments by the \textit{fermata}\footnote{A \textit{fermata} represents a full-stop marker in a lyric in chorales.} positions and treated the segments as independent sequences. We trained with three random seeds: 123, 456, and 789. The minibatch size was 8.

Other settings were the same between JSBChorales60 and JSBChorales371. The number of training epochs was 480 for phase 1 and 240 for phase 2. The best model was chosen by the negative log-likelihood on the development data. The training was stopped prematurely if the best model was not updated in 80 consecutive epochs. The optimizer was Adam\cite{kingma2015}, and the learning rate was 1e-3.

\subsection{Automatic evaluation results of harmonic analysis}
\label{sec:results}
\begin{table}[hbtp]
\centering
\small
\caption{Evaluation results of \textit{Accuracy} on JSBChorales60.}
\label{tab:result-harmony-analysis-60}
\begin{tabular}{l|c|c|c}
\hline
model & method & Full Chord & Root Chord \\ \hline
HMPerceptron\cite{radicioni2010} & supervised & 80.1 & - \\
HMPerceptron (re-experimented\cite{masada_crf}) & supervised & 77.2 & 84.8 \\
Semi-CRF\cite{masada_crf} & supervised & 83.2 & 88.9 \\ \hline
Melisma\cite{temperley_web} (reported by\cite{masada_crf}) & rule-based & - & 84.3 \\ \hline
Ours & unsupervised & 66.8 & 79.2 \\
\hline
\end{tabular}
\end{table}
The \textit{Accuracy}\footnote{\textit{Accuracy} is the percentage of correctly predicted labels from the total number of events in the dataset.} of our model on the JSBChorales60 dataset is shown in 
Table~\ref{tab:result-harmony-analysis-60}. Note that the proposed model did not consider enharmonic notes. 
Therefore, we converted the root names to pitch classes when evaluating our model.
In this sense, the evaluation is not under exactly the same conditions as the other models in the table.
However, as mentioned earlier, we do not consider the enharmonic issue a limitation since it can be resolved if the information in the original score remains.

Table~\ref{tab:result-harmony-analysis-60} shows that our model underperformed compared to supervised learning and sophisticated rule-based models.
However, we have contributed to the advancement of harmonic analysis with unsupervised learning by enabling unsupervised learning of model parameters and presenting evaluation scores to demonstrate the current performance of unsupervised learning.

\begin{table}[hbtp]
\centering
\small
\caption{Evaluation results of \textit{Accuracy} on JSBChorales371.}
\label{tab:result-harmony-analysis-371}
\begin{tabular}{l|c|c|c}
\hline
model & Key & Full RN & Root RN \\ \hline
Ours & 74.2 & 61.6 & 66.9 \\
\hline
\end{tabular}
\end{table}
As mentioned in Section~\ref{sec:setups}, we evaluated chord names on JSBChorales60, whereas we can evaluate harmonic analysis using Roman numerals (RNs) on JSBChorale371.
Analysis with Roman numerals requires simultaneous recognition of the keys and roots; therefore, it is more complex than recognizing chord symbols.
The \textit{Accuracy} scores for JSBChorales371 with our model are shown in Table~\ref{tab:result-harmony-analysis-371}\footnote{The human annotation sometimes gives multiple interpretations of a single chord at the start or end of modulation, but in this evaluation, the last label was used.}.
To the best of our knowledge, this is the first report comparing the manual and unsupervised harmonic analyses attached to the Riemenschneider numbers 1--20 of J. S. Bach's chorales\footnote{$\rightarrow$ footnote~\ref{footnote:chorale-analysis}}.
However, our model still has challenges, as seen in the scores obtained.
In the next section, we will discuss these issues by presenting the results of the analysis.

\subsection{Discussion on examples of the obtained analyses}
\begin{figure}[htbp]
\centering
\includegraphics[width=\linewidth]{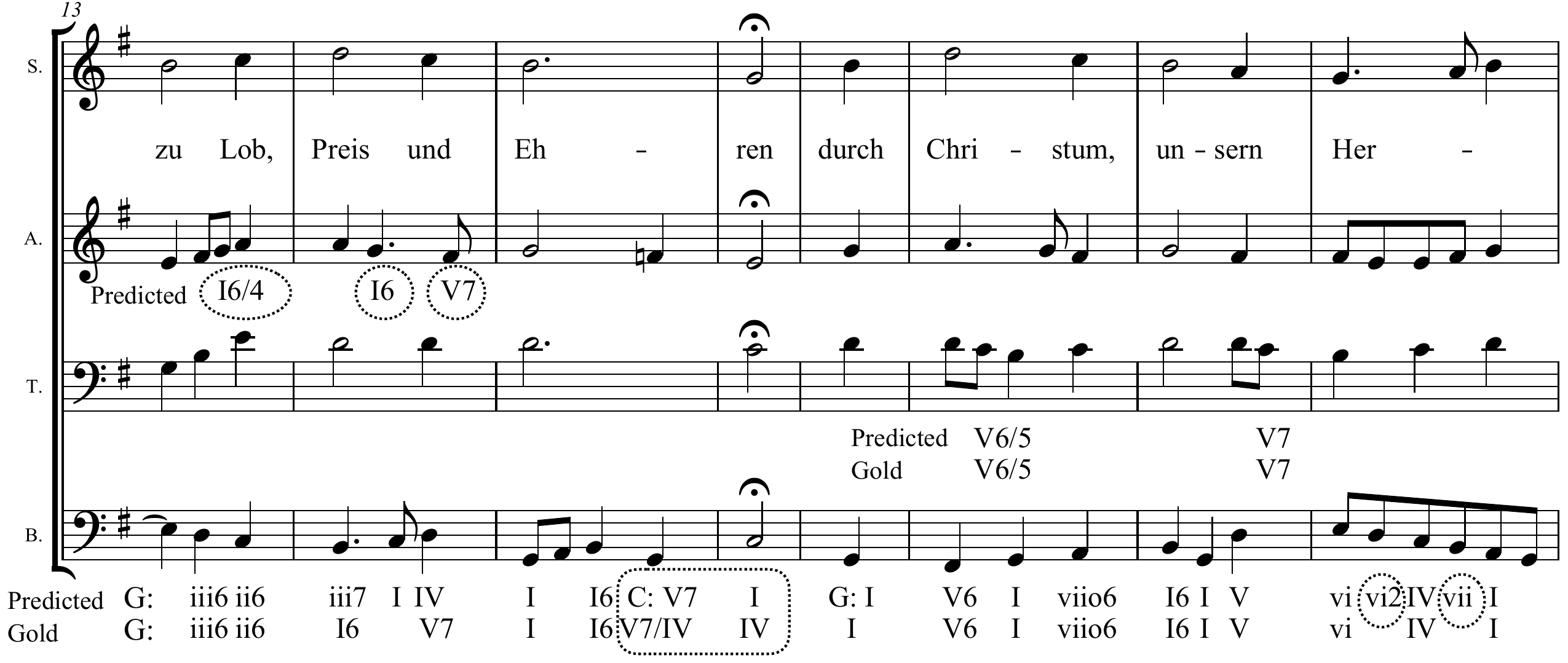}
\caption{Harmonic analysis of BWV269 (Riemenschneider No.1), bars 13 -- 20.}
\label{fig:riemen1}
\end{figure}
Figure~\ref{fig:riemen1} shows the harmonic analyses of BWV269, bars 13--20.
The labels are displayed below the bass notes, if present, or at the places where the note exists, in preference of lower parts.
A typical error in our model, exemplified by dotted circles in Figure~\ref{fig:riemen1}, is the extra annotation of passing tones.
As shown in Figure~\ref{fig:riemen1}, the gold annotation distinguished V6--V6/5 (bar 18) and V--V7 (bar 19), while it did not distinguish vi--vi2 (bar 20), unlike our model's prediction.
The difference between the V and vi cases may be suggested by whether they were on the Bass passing note or not; however, these differences were difficult to detect in our model, which is based on statistical unsupervised learning.

Another issue with our model is that it does not support borrowed chords; therefore, as shown by the dotted square (bars 15 and 16) in Figure~\ref{fig:riemen1}, borrowed chords are sometimes detected as modulations.

\begin{figure}[htbp]
\centering
\includegraphics[width=\linewidth]{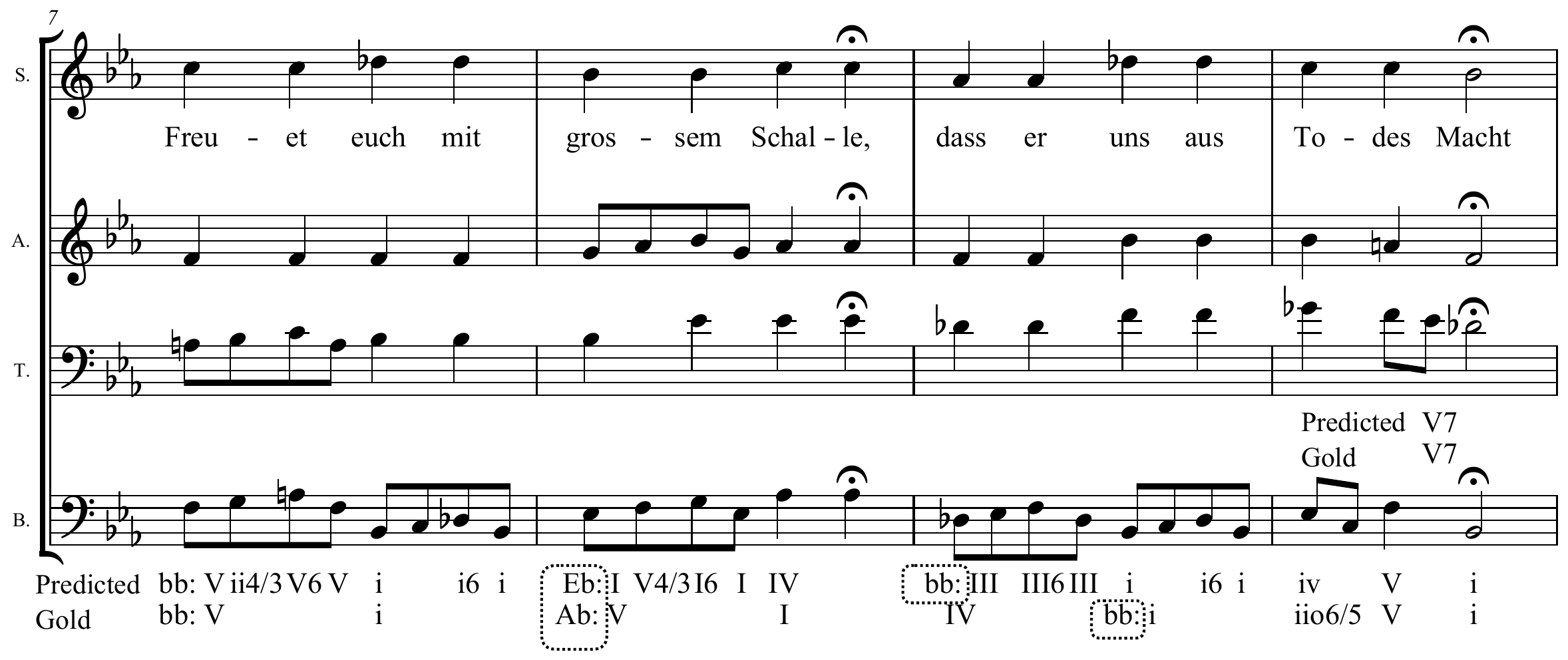}
\caption{Harmonic analysis of BWV40.8 (Riemenschneider No.8), bars 7 -- 10.}
\label{fig:riemen8}
\end{figure}
The process of identifying keys is more complex than commonly thought.
Figure~\ref{fig:riemen8} shows an example.
As shown by dotted circles, both gold and predicted analyses detect a modulation at bar 8.
However, a consistent four-degree ascending root progression (F $\rightarrow$ Bb $\rightarrow$ Eb $\rightarrow$ Ab $\rightarrow$ Db) can be observed in bars 7--9.
Therefore, it is possible to interpret the progression as (V $\rightarrow$ i $\rightarrow$ IV $\rightarrow$ VII $\rightarrow$ III) without any modulation.

\subsection{Tonality derived from the learned root-transition probabilities}
\label{sec:tonic}
As noted earlier, we did not assume that the two modes we set were major and minor, nor did any assumption to tonic notes.
Many works have considered that the pitch class with the highest percentage of notes is the tonic\cite{krumhansl2001cognitive,temperley_kru,wang2012,hu2009}.
We argue, however, that the tonic should be recognized in the context of harmonic progressions.
In particular, we consider that tonic is the largest component of the stationary distribution of the root-transition probability matrix.
The stationary distribution of a Markov chain is also used in tasks such as measuring the importance of a Web site\cite{page1998pagerank}.

Here, we briefly describe the stationary distribution of a Markov chain\cite{serfozo2009basics}.
If the initial state vector is $\bm{\pi}_0$ and the transition probability matrix is $M$, the state probability vector after the $t$ steps is as follows.
\begin{align}
\bm{\pi}_t^\top &= \bm{\pi}^\top_0 M^{t}
\end{align}
The following equation holds if $\bm{\pi}_t$ approaches a constant value $\bm{\pi}$ when $t \rightarrow \infty$; this $\bm{\pi}$ is called the stationary distribution in a Markov chain.
\begin{align}
\label{eq:stationary-distribution}
\bm{\pi}^\top &= \bm{\pi}^\top M,~~\sum_l \pi_l = 1
\end{align}
The equation (\ref{eq:stationary-distribution}) means that the stationary distribution is an eigenvector for eigenvalue 1 of the matrix $M$.

In our study, the matrix of root transition probabilities corresponds to $M$ in (\ref{eq:stationary-distribution}).
However, since our model is a semi-Markov model and the root transition probabilities do not include self-transitions, we modify the root-transition matrix by combining it with the inverse of the average duration probability $a^\textrm{dur.}$ to construct $M$.
\begin{align}
\label{eq:comb-root-transition}
(a^\textrm{dur.})^{-1} &= \frac{1}{\sum^{N_d - 1}_{d=0} (d + 1) \pi^{z_d}_d} \\
M_m &= (1 - (a^\textrm{dur.})^{-1}) \bm{I} + (a^\textrm{dur.})^{-1}
\begin{pmatrix}
\pi_{j=0|i=0,m} \dots \pi_{j=12|i=0,m} \\ 
\ddots \\
\pi_{j=0|i=12,m} \dots \pi_{j=12|i=12,m}
\end{pmatrix}
\end{align}
$\pi^{z_d}_d$ is the same duration probability in eq. (\ref{eq:duration-prob}), and $\pi_{j|i,m}$ are root-transition probabilities (\ref{eq:root-transition-prob}).

\begin{figure}[htbp]
\centering
\includegraphics[width=10cm]{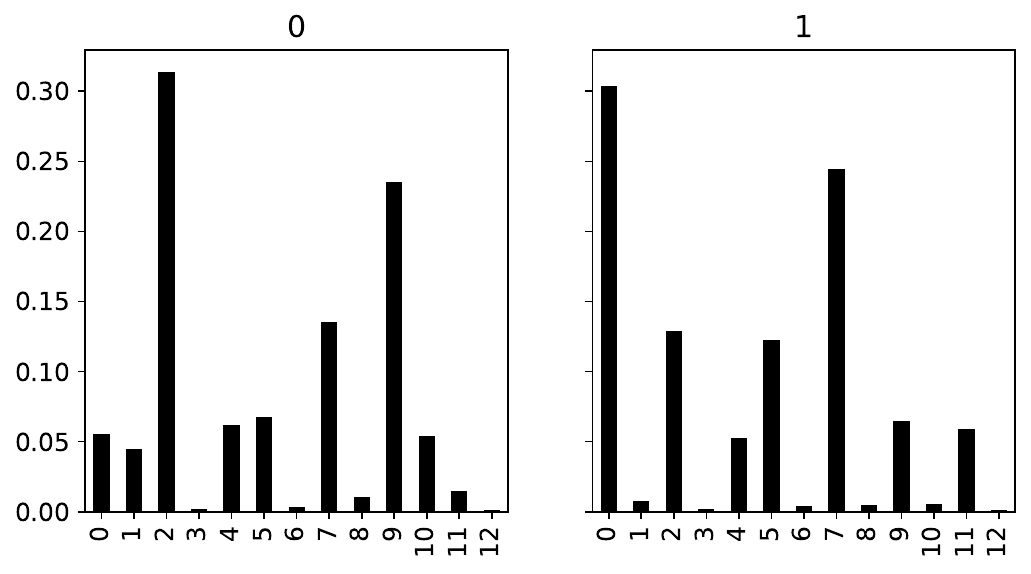}
\caption{Result of the stationary distribution of root pitch classes (JSBChorales371, seed=789). The left side is mode $m = 0$, and the right is mode $m = 1$.}
\label{fig:stationary}
\end{figure}
The obtained stationary distributions are shown in Figure~\ref{fig:stationary}.
In mode 0, the pitch class with the highest percentage was 2, which made D the tonic. On the other hand, in mode 1, C was the tonic.
Thus, $m=0$ can be interpreted as the d-minor key and $m=1$ as the C-major key.
The reason why the d-minor was learned instead of the a-minor can be interpreted as follows.
Several pieces in the JSBChorales371 dataset were written in the Dorian mode, which can be identified by having one less key signature than modern notation.
Hence, in the Phase 1 training (Section~\ref{sec:training}), the learned mode seemed to be a mixture of d-minor and a-minor.
After that, mode 0 converged into d-minor in the Phase 2 training.
The learned mode 0 had nearly equal proportions of 0 and 1 pitch classes corresponding to degree VII. Thus, the proposed model learned the ambiguity of degree VII in the minor key.
It is also interesting to note that the importance of I and V was similar for both modes, but in d-minor, IV was more important than II.

Furthermore, the pitch class probability can be obtained using the stationary distribution, and the logits of the pitch class given a root computed in eq. (\ref{eq:marginal-emission-logit}).
\begin{align}
p(pc | m, r) &= \textrm{sigmoid}(v_{pc|r,m}) \\
p(pc | m) &= \sum_r p(pc | m, r) p(r)
\end{align}
The $p(r)$ is the root probability calculated as the stationary distribution of the root-transition matrix described above. 

\begin{figure}[htbp]
\centering
\includegraphics[width=10cm]{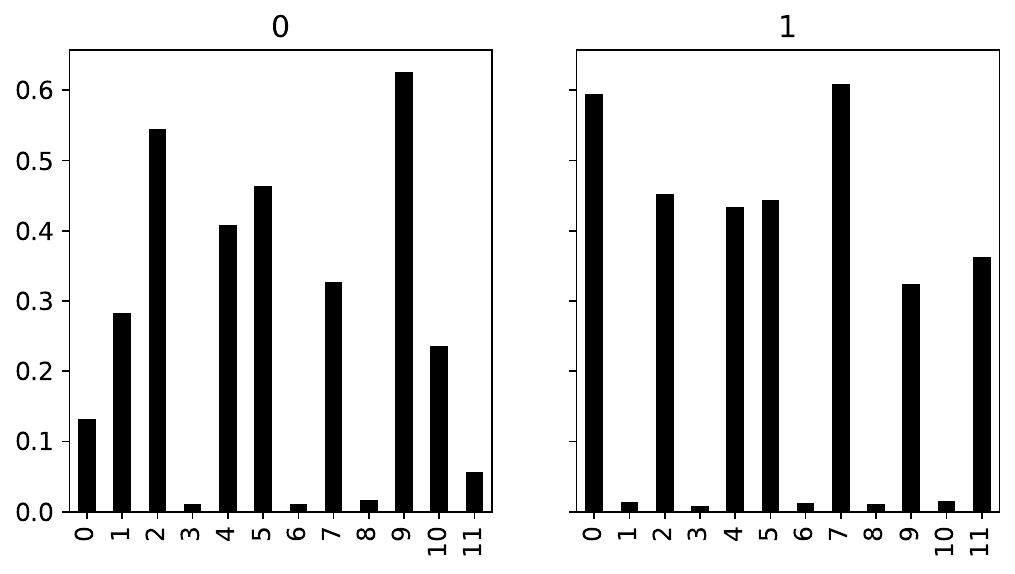}
\caption{Result of the pitch class probabilities (JSBChorales371, seed=789). The left side is mode $m = 0$, and the right is mode $m = 1$.}
\label{fig:pitch-class}
\end{figure}

Obtained pitch class probabilities are shown in Figure~\ref{fig:pitch-class}.
Unlike the \textit{Key Profiles}\cite{krumhansl2001cognitive,temperley_kru,hu2009}, the pitch class probability of V could be larger than I. 
This may be partly due to the output representation as a binary vector of pitch classes. 
However, the stationary distribution in Figure~\ref{fig:stationary} appears to express the importance of each degree more clearly than the pitch class probabilities in Figure~\ref{fig:pitch-class}.

\section{Conclusion}
\label{sec:conclusion}
This paper proposed an unsupervised harmonic analysis based on the neural hidden semi-Markov model (HSMM). 
The model was constructed with neural networks that approximate probability distributions in the HSMM.
This technique allowed feasible unsupervised learning of model parameters, which has been difficult in previous studies.
In addition, we introduced chord quality templates, which enabled the harmonic analysis with known chord labels such as chord names and Roman numerals. 

Although we presupposed that the number of modes was two and keys were equivalent if transposed, we did not make any other assumption for the two modes. Nevertheless, our model could find the minor and major modes and their tonic notes properly, as was discussed in Section ~\ref{sec:tonic}.

However, the \textit{Accuracy} results on labeled data and examples of obtained analyses suggest that our model still has room to improve.
An important future work is to distinguish passing tones.
Since our model is a generative model, it assigns probability to all notes in a score.
This may cause excess labeling of passing notes, which would be more severe for instrumental music.
We may consider a method of probabilistic estimation of passing notes and changing the generation mechanism depending on whether they are passing notes or not.

If the issue of passing tones is resolved, the proposed method should apply to a wider range of music.
The true value of unsupervised learning would be demonstrated for music after the late Romantic period, for which there is little labeled data.
Furthermore, the model may be effective for pre-Renaissance music since our model could learn modes and tonics unsupervised.

\begin{small}
\section*{Acknowledgments}
This research has been supported by JSPS KAKENHI No. 23K20011.
Computational resource of AI Bridging Cloud Infrastructure (ABCI) provided by National Institute of Advanced Industrial Science and Technology (AIST) was used. 
\end{small}

\begin{small}
\bibliographystyle{abbrv}
\bibliography{references.bib}
\end{small}

\end{document}